# Ophthalmic Biomarker Detection with Parallel Prediction of Transformer and Convolutional Architecture

Md. Touhidul Islam[1], Md. Abtahi Majeed Chowdhury[2], Mahmudul Hasan[3], Asif Quadir[4], Dr. Lutfa Aktar[5]
Department of Electrical and Electronic Engineering
Bangladesh University of Engineering and Technology
Dhaka-1000, Bangladesh
[1]touhid2.718@gmail.com, [2]1806106@eee.buet.ac.bd, [3]hasanmahmudul11085@gmail.com,
[4]asif.quadir187@gmail.com , [5]lutfaakter@eee.buet.ac.bd

*Abstract*—**Ophthalmic diseases represent a significant global health issue, necessitating the use of advanced precise diagnostic tools. Optical Coherence Tomography (OCT) imagery which offers high-resolution cross-sectional images of the retina has become a pivotal imaging modality in ophthalmology. Traditionally physicians have manually detected various diseases and biomarkers from such diagnostic imagery. In recent times, deep learning techniques have been extensively used for medical diagnostic tasks enabling fast and precise diagnosis. This paper presents a novel approach for ophthalmic biomarker detection using an ensemble of Convolutional Neural Network (CNN) and Vision Transformer. While CNNs are good for feature extraction within the local context of the image, transformers are known for their ability to extract features from the global context of the image. Using an ensemble of both techniques allows us to harness the best of both worlds. Our method has been implemented on the OLIVES dataset to detect 6 major biomarkers from the OCT images and shows significant improvement of the macro averaged F1 score on the dataset.**

I. INTRODUCTION

In the field of Medical Image Analysis, deep learning has ushered in new possibilities for ophthalmic biomarker detection. While CNN has been the dominant architecture for image processing for quite some time, recently vision transformers are showing performance competitive with CNN. CNNs can find features on local context as they use convolutional layers that slide small filters over the input data. Vision transformers, on the other hand, can find features in a global context through self-attention mechanism. However, Transformer models tend to be very big to provide superior performance. Training and inferring with such huge models are both resourceful and time-consuming. In this paper, we have shown a novel approach for harnessing the good of both convolutional and transformer world that can provide superior results with relatively smaller models. We have used a small transformer model with about 100M parameters and a small CNN model with about 55M parameters. Both models are small, compared to the current SOTA models, and their individual performances are below the current SOTA accuracies too. However, their ensemble has shown significant improvement from their individual performances. Also, the overall model is fast to train because of the small number of parameters and the individual models' fast training characteristics.

II. RELATED WORK

There have been many works in the field of CNN and vision transformer and their use in medical diagnosis. The OLIVES dataset we worked on, however, provides scope of several possible approaches. A significant portion of the dataset consists of unlabeled images. Of the 78108 OCT scans available in the dataset, only 9408 images are biomarker labeled. As a result, self-supervised learning is a viable option. Also, All the OCT images have clinical labels for BCVA and CST. Hence, a clinically labeled self-supervised learning with contrastive loss approach was proposed in [2]. However, the self-supervised approach suffers from the problem that it cannot utilize the image label of 9408 OCT images.

The dataset also provides fundus of the eyes along with their OCT. As a result, multimodal learning is another possible option with the 3 available modalities i.e., OCT scan, fundus, and clinical label. However, medical datasets with such multiple modalities are less prevalent [1], and hence work on this approach is comparatively limited. In [3], authors have presented a multimodal approach with OCT, colored fundus image, and relevant patient data as 3 modalities.

In [4], the authors did a comparative study on CNN and vision transformers for medical image diagnosis. Although vision transformers cannot surpass CNNs by a big margin, they have shown comparable performance with CNN.

The ensemble of multiple weak deep learning models to produce a stronger model together has been a renowned technique for improving deep learning model performance. In [4], authors showed a comparative study of several ensemble techniques like stacking, augmenting, and bagging on 4 medical image datasets.

III. PROPOSED APPROACH

*A. Data Preprocessing*

Most of the OCT images contain a significant amount of noise. However, removing the noise from the OCT images is tricky as the key features for two diseases (namely Partially Attached Vitreous Face (PAVF) and Vitreous debris (VD)) are very much noise-like which is apparent from figure 1. As a result, applying aggressive noise removal techniques may take away some key features off the image. We applied a simple brightness and contrast adjustment method to reduce the noise of the images. The brightness of the image was decreased by a factor and contrast was increased by another factor. Increasing brightness corresponds to adding a constant value to the pixels and adjusting contrast corresponds to multiplying a constant with the pixels. Hence we essentially used a linear transformation on the pixel to remove the noise to a reasonable degree. The degree of noise removal was essential, as too aggressive noise removal can take away features. The factors of the linear transformation were thus experimentally determined to give the best possible result.

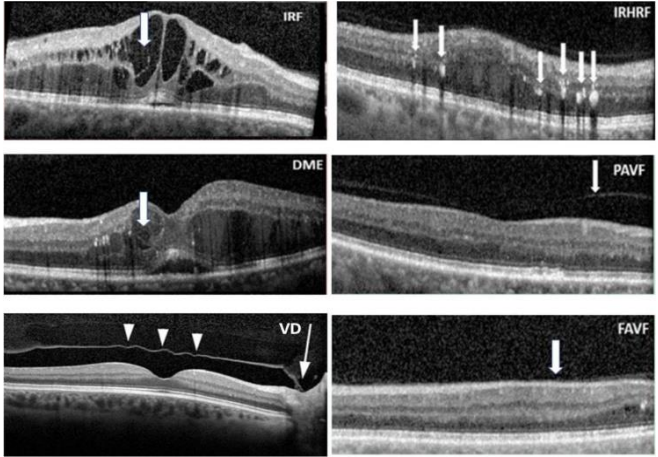

Fig 1. The six different biomarkers studied in this experiment with their key diagnostic features highlighted.

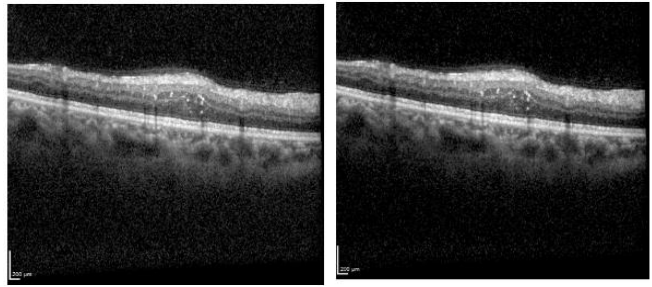

Fig 2. OCT image before and after noise removal (Left: Original Image, right: noise removed)

### B. Data Augmentation

- Random Crop has been taken form dataset images.
- Gaussian Blur is also taken into consideration.
- Flipping the images horizontally and Random Perspective and affine has also been utilized.
- As most of the OCT scan background is black in color, we have transformed the image of white background after detecting the background.
- For detecting background we have threshold cutoff method and replacing an average black background.

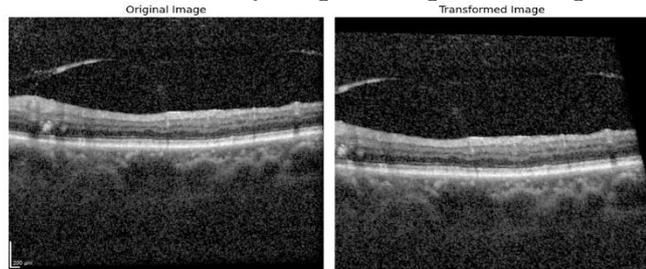

Fig. Original Image(Left) and after applying random perspective

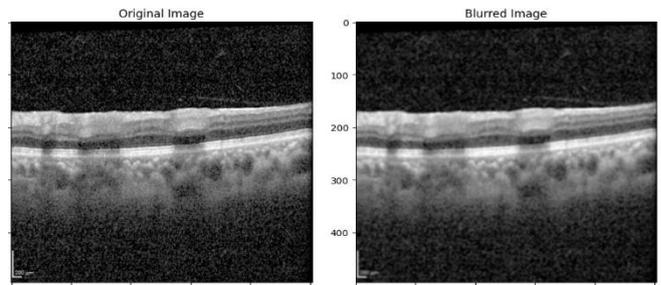

Fig. Before and after applying Random Gaussian Blur

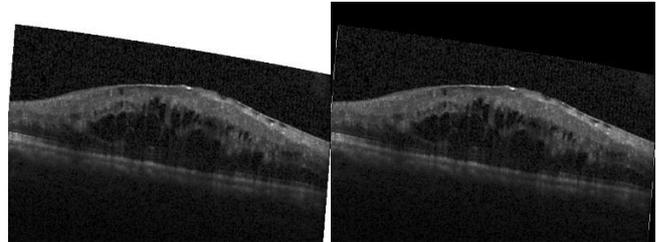

Fig. After detecting Background and transforming it to black

### C. EfficientNet

EfficientNet[12][13] is a CNN based model that uniformly scales all dimensions of depth/width/resolution using a compound coefficient. Efficientnet can train really fast while providing decent performance. There are several EfficientNet architectures available. We tested EfficientNet-B7, EfficientNetV2-M, and EfficientNetV2-L on our dataset. We found EfficientNetV2-M best performing for our task.

Because, the decision of availability of a disease depends not on the whole image but on some specific key features of the image(as indicated in figure 3), giving attention to particular portion of the image is beneficial[14][15]. EfficientNetV2 integrates attention block like Squeeze and Attention block in the architecture as indicated on figure 3. As a result, EfficientNetV2 has performed better than the basic EfficientNet-B7.

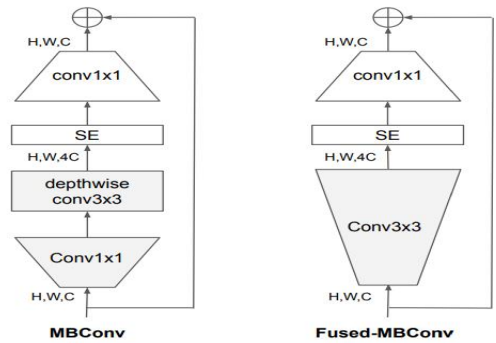

Fig 3. EfficientNetV2 MBConv and Fused-MBConv block

### D. MAX-VIT

MaxViT[6][11] is basically a family of hybrid (CNN+ViT) image classification models. It achieves better performance than CNN models, or ViT models considering both parameters and Flops efficiency.

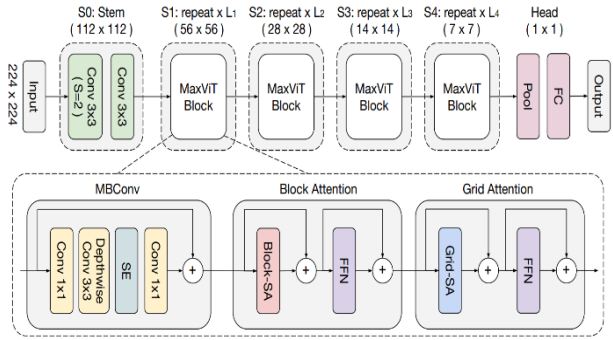

Fig 4. Architecture of MaxVIT

MaxViT model contains a new type of attention module named multi-axis self-attention (Max-SA).

In Fig.5, It decomposes the fully dense attention mechanisms into two sparse forms – window attention and grid attention.

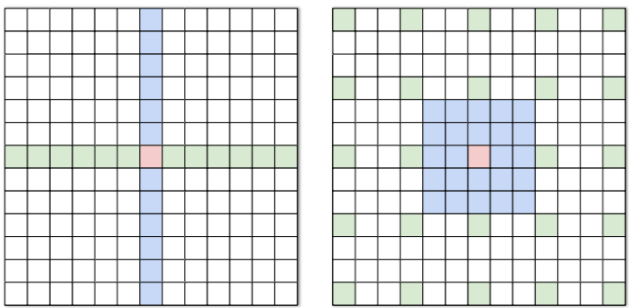

Fig. 5 multi axis self attention

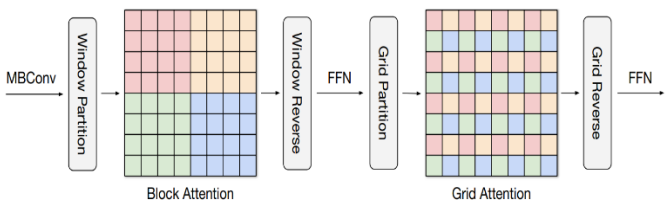

Fig. 6 global and local context in MaxViT

This reduces the quadratic complexity of vanilla attention to linear without losing non-locality. For this reason, MaxViT is capable of performing both local and global spatial interactions[12]. ViT models take much longer time to train, while MaxViT takes less time than that. To train ViT models properly, we need more computational resources. MaxViT needs fewer computational resources in that case. If MaxVit is trained improperly, it could show bias or fairness issues.
In our context of Ophthalmic Biomarker detection, diseases depend on the specific key features in the images. Giving attention to specific key featured areas of the images is our main concern. For this, we have to make sure that we do not want to lose any global or local information in any stage. MaxViT outperforms any other model for this specific task.

### E. ENSEMBLE

*1) Necessity of ensemble technique:*
Ensemble technique[5][7] is a great method to work with for a stable prediction model. In learning models noise, variance and bias are the major sources of error. The ensemble methods in machine learning help minimize these error-causing factors, thereby ensuring the accuracy and stability of machine learning algorithms. In our research with OCT data, it is observed that OCT scans can vary a lot for different scanning mechanisms, brightness, orientation and unavoidable possibility of environmental interference[8]. To work with such huge yet varying data, relying on a single pretrained model doesn't performs well. In our approach we have deployed combinations of transformer based models and CNN-architecture based models[9][10].

*2) Combined Prediction model with MAXVIT and Efficient-NET:*
Combining CNN-architecture based model and Transformer-based model is a noble approach to deal with medical classification model[17][18]. As seen in our research, Transformer based model can't single handedly outperform CNN-based architecture. Convolution layers works fine in extracting global feature of an image[19] whereas an attention layer can focus local feature of data and predict accordingly[20]. We have bought together the strength of both architecture and proposed a reliable combination through ensemble.

*3) Paralled Model on different Dataset:*
We have observed that OCT scans from different sources can vary a lot depending on the manual ways of handling the scanner machine. It can be tilted as before. So, we created 5 paralled branch:
- Both Trex & Prime dataset is included in training EfficientNet(M)
- Both Trex & Prime dataset is used to train MAXVIT.
- Only Trex is used to train both EfficientNet and MAXVIT
- Only Prime is used to train EfficientNet

The main idea behind this is that now for different kind of test data will go through all 5 branches and give separate bias result. It then go through weighted average and rounding process and final is result is produced.

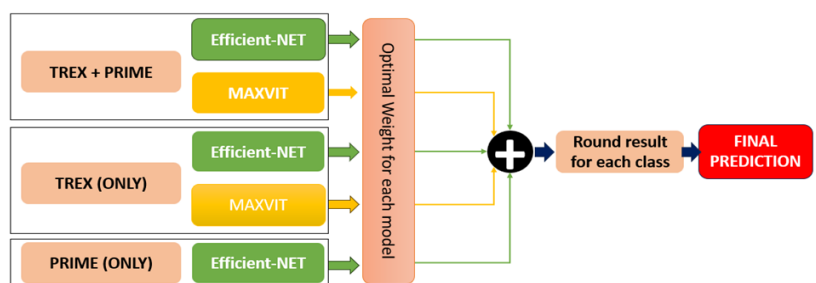

Figure 7: Overall ensemble architecture

*4) Selection of Optimal Weight for each Parallel branch:*
To Select Optimal Weight for each branch we have created a validation set which is a mixture of images form TREX & PRIME. Then, we took "Iterative approach" to find the best suited weights. We have done it on multiple validation set and finally it is seen that the best result doesn't come for a single model but rather a distributed weight gives the best result.

## IV. DATASET

The modified version of the OLIVES (Ophthalmic Labels for Investigating Visual Eye Semantics) dataset was used to perform our experiments. This dataset is composed of 78108 OCT scans from two clinical trials. Every image is associated with the clinical information of Eye identity, BCVA, and CST that was collected during the process of patient treatment during the clinical trial. Fundus image of the associated OCT volume was also provided. The modified OLIVES dataset provided contained 6 biomarkers information graded by professional graders for 9408 OCT images. The biomarkers are Intraretinal Hyperreflective Foci (IRHRF), Partially Attached Vitreous Face (PAVF), Fully Attached Vitreous Face (FAVF), Intraretinal Fluid (IRF), and Diffuse Retinal Thickening or Diabetic Macular Edema (DRT/ME) and Vitreous debris (VD). In total, the OLIVES dataset provides data from 96 unique eyes.

The dataset was divided into a training set and a test set. For validation purpose, a portion of the train set was used. If the total pool of images is randomly split, it would create overfit problem. This is because 2 OCT slices of the same volume are very similar in appearance and diagnosis. Hence, Eye eye-wise random split was used.

## V. EXPERIMENTS AND RESULTS

Table showing the individual result produced by EfficientNet on the first 70% of test dataset:

| Model | Number of parameters | Image Modality | Macro Average F1 Score |
|---|---|---|---|
| EfficientNet-B7 | 66M | OCT Slice | 0.672 |
| EfficientNet-V2-L | 120M | OCT Slice | 0.709 |
| EfficientNet-V2-M | 54M | OCT Slice | 0.715 |

As apparent from the result, we were able to fine tune the smaller model to provide better accuracy.

Individual result produced by MAXVIT on the first 70% of test dataset:

| Model | Number of parameters | Image Modality | Macro Average F1 Score |
|---|---|---|---|
| MaxViT-base | 119M | OCT Slice | 0.703 |

Weighted Distribution which gave the best result:

| Parallel Branch Model | Dataset | Weight |
|---|---|---|
| EfficientNet-V2-M | Trex + Prime | 0.1 |
| Maxvit-base | Trex + Prime | 0.45 |
| EfficientNet-V2-M | Trex | 0.1 |
| MaxViT-base | Trex | 0.25 |
| EfficientNet-V2-M | Prime | 0.1 |

The ensembled model was tested upon the 2nd test set provided:

| Model | Test set | Personalized F1 score |
|---|---|---|
| **Ensembled | VIP Cup 2023 – Phase 2 | 0.8116 |

## VI. CONCLUSION

Due to high granularity of 3 biomarkers (IRHRF, PAVF, FAVF), the task of detecting biomarkers was not easy. Moreover, modern SOTA architectures are very big, resource-hungry, and time-consuming. By ensembling EfficientNetV2 and MaxVit we were able to exploit both local context features and global context features. Hence the overall model provides superior accuracy even with smaller individual models. As the model is small, training and inference is fast. Such fast and small models can be rapidly trained and adapted for other medical image classification tasks too.

## ACKNOWLEDGMENT

We want to thank the organizers of VIP Cup 2023 as well as IEEE for launching a competition with a problem that is really important in the context of our time.